\documentclass[11pt,a4paper]{article}
\usepackage[hyperref]{emnlp2018}
\usepackage{times}
\usepackage{latexsym}
\usepackage{enumitem}
\usepackage{multirow}
\usepackage{graphicx}
\usepackage{url}
\usepackage[title]{appendix}

\usepackage{amsmath}

\usepackage{fancyhdr}
\usepackage{geometry}
\aclfinalcopy 

\title{Word Sense Induction with Neural biLM and Symmetric Patterns}

\author{Asaf Amrami$^{~\dagger}$ \and Yoav Goldberg$^{~\dagger~\ddagger}$ \\[0.5em]
  $\dagger$ Computer Science Department, Bar Ilan University, Israel \\
  $\ddagger$ Allen Institute for Artificial Intelligence \\[0.5em]
  {\tt \{asaf.amrami, yoav.goldberg\}@gmail.com}
  }

\date{26.8}

\begin{document}
\maketitle
\begin{abstract}

  An established method for Word Sense Induction (WSI) uses a language model
  to predict probable substitutes for target words, and induces senses by clustering these resulting substitute vectors.
  
  We replace the ngram-based language model (LM) with a recurrent one. Beyond being more accurate, the use of the recurrent LM allows us to effectively query it in a creative way, using what we call \emph{dynamic symmetric patterns}.
  The combination of the RNN-LM and the dynamic symmetric patterns results in strong substitute vectors for WSI, allowing to surpass the current state-of-the-art on the SemEval 2013 WSI shared task by a large margin.

\end{abstract}

\section{Introduction}

We deal with the problem of \emph{word sense induction} (WSI): given a target lemma and
a collection of within-sentence usages it, cluster the usages (\textbf{instances})
according to the different senses of the target lemma. For example, for the
sentences:
\begin{enumerate}[label=(\alph*)]
\item  We spotted a large \textbf{\textit{bass}} in the ocean. 
\item The \textbf{\textit{bass}} player did not receive the acknowledgment she deserves.
\item  The black sea \textbf{\textit{bass}}, is a member of the wreckfish family.
\end{enumerate}
We would like to cluster (a) and (c) in one group and (b) in
another.\footnote{This example shows \emph{homonymy}, a case where the same word
form has two distinct meaning.  A more subtle case is \emph{polysemy}, where
the senses share some semantic similarity. In ``She played a low
\textbf\emph{bass} note'', the sense of \emph{bass} is related to the sense in
(b), but distinct from it. The WSI task we tackle in this work deals with both
cases.}
Note that some mentions are ambiguous. For example, (d) matches both
the music and the fish senses:
\begin{enumerate}[resume,label=(\alph*)]
    \item \textbf{\emph{Bass}} scales are the worst.
\end{enumerate}
This calls for a \emph{soft clustering}, allowing to probabilistically associate a given mention
to two senses.

The problem of WSI has been extensively studied with a series of shared tasks on
the topic \cite{semeval2007wsi,semeval2010wsi,semeval2013wsi}, the latest being
SemEval 2013 Task 13 \cite{semeval2013wsi}.
Recent state-of-the-art approaches to WSI rely on generative graphical models
\citep{lau2013unimelb,wang2015sense,komninos2016structured}.
In these works, the sense is modeled as a latent variable that influences the context of the
target word. The later models explicitly differentiate between local (syntactic, close
to the disambiguated word) and global (thematic, semantic) context features.

\paragraph{Substitute Vectors}
\citet{baskaya2013ai} take a different approach to the problem, based on
\emph{substitute vectors}. They represent each instance as a distribution
of possible substitute words, as determined by a language model (LM). The substitute vectors are then clustered to obtain senses.
 
 \citet{baskaya2013ai}
derive their probabilities from a 4-gram language model. Their system (AI-KU) was one of the best performing at the time of SemEval 2013 shared task. 
Our method is inspired by the AI-KU use of substitution based sense induction, but deviate from it by moving to a recurrent language model. Besides being more accurate, this allows us to further improve the quality of the derived substitutions by the incorporation of \emph{dynamic symmetric patterns.}

\paragraph{BiLM}
Bidirectional RNNs were shown to be effective
for word-sense disambiguation and lexical substitution tasks
\cite{context2vec,yuan2016semi,raganato2017neural}. We adopt the ELMo biLM model of \citet{peters2018deep}, which was shown to produce very competitive results for many NLP tasks. 
We use the pre-trained ELMo biLM provided by \citet{peters2018deep}.\footnote{We thank the ELMo team for sharing the pre-trained models.} However, rather than using the LSTM state vectors as suggested in the ELMo paper, we opt instead to use the predicted word probabilities. Moving from continuous and opaque state vectors to discrete and transparent word distributions allows far better control of the resulting representations (e.g. by sampling, re-weighting and lemmatizing the words) as well as better debugging opportunities.

As expected, the move to the neural biLM already outperforms the AI-KU
system, and matches the previous state-of-the-art. However, we observe that the substitute vectors do not take into account the
disambiguated word itself. We find that this often results in noisy substitutions.
As a motivating example, consider the sentence ``the doctor recommends
\emph{oranges} for your health''. Here, \emph{running} is a perfectly good substitution, as the ``fruitness''
of the target word itself isn't represented in the context. 
We would like the substitutes
word distribution representing the target word to take both kinds of information---the context as well as the target word---into account.

\paragraph{Dynamic Symmetric Patterns}
Our main proposal incorporates such information. It is motivated by
Hearst patterns
\cite{hearst1992automatic,Widdows:2002:GMU:1072228.1072342,schwartz2015symmetric},
and made possible by neural LMs.
Neural LMs are better in capturing long-range dependencies, and can handle and predict unseen text by generalizing from similar contexts.
Conjunctions, and in particular the word \emph{and}, are known to combine expressions of the same kind.  Recently, \citet{schwartz2015symmetric} used conjunctive symmetric patterns
to derive word embeddings that excel at capturing word similarity. Similarly, \citet{kozareva2008semantic} search for doubly-anchored patterns including the word \emph{and} in a large web-corpus to improve semantic-class induction. The method of \citet{schwartz2015symmetric} result in context-independent embeddings, while that of \citet{kozareva2008semantic} takes some context into account but is restricted to exact corpus matches and thus suffers a lot from sparsity.

We make use of the rich sequence representation capabilities of the neural biLM to derive \emph{context-dependent
symmetric pattern substitutions}.
Relying on the generalization properties of neural language models and the abundance of the ``X and Y'' pattern, we present the language model with a dynamically created incomplete pattern, and ask it to predict probable completion candidates. Rather than predicting the word
distribution following \emph{the doctor recommends \underline{\hspace{1em}}}, we instead
predict the distribution following \emph{the doctor recommends oranges and
\underline{\hspace{1em}}}. This provides substantial improvement, resulting in
state-of-the-art performance on the SemEval 2013 shared task.

The code for reproducing the experiments and our analyses is available at \url{https://github.com/asafamr/SymPatternWSI}.

\section{Method}

Given a target word (lemma and its part-of-speech pair), together with several sentences in which the target word is used (instances), our goal is to cluster the word usages such that each cluster corresponds to a different sense of the target word.
Following the SemEval 2013 shared task and motivating example (d) from the
introduction, we seek a soft (probabilistic) clustering, in which each word instance is assigned with a probability of belonging to each of the sense-clusters.

Our algorithm works in three stages:
(1) We first associate each instance with a probability distribution over in-context word-substitutes. This probability distribution is based on a neural biLM (section \ref{ssec:bilm}). 
(2) We associate each instance with $k$ representatives, each containing multiple samples from its associated word distributions (section \ref{ssec:sampling}). 
(3) Finally, we cluster the representatives and use the hard clustering to derive a soft-clustering over the instances (section \ref{ssec:clustering}).

We use the pre-trained neural biLM as a black-box, but use linguistically motivated processing of both its input and its output: we rely on the generalization power of the biLM and query it using \emph{dynamic symmetric patterns} (section \ref{ssec:sympat}); and we lemmatize the resulting word distributions.

\paragraph{Running example}
In what follows, we demonstrate the algorithm using a running example of inducing senses from the word \textit{sound}, focusing on the instance sentence:

\noindent\fbox{\parbox{\columnwidth}{
\emph{I liked the \textbf{\textit{sound}} of the harpsichord.}}}

\subsection{biLM Derived Substitutions}
\label{ssec:bilm}

We follow the ELMo biLM approach \cite{peters2018deep} and consider two separately trained
language models, a forward model trained for predicting
$p_{\rightarrow}(w_i|w_1,...,w_{i-1})$ and a backward model
$p_{\leftarrow}(w_i|w_n,...,w_{i+1})$. Rather than combining the two models' predictions into a
single distribution, we simply associate the target word with two distributions,
one from $p_{\rightarrow}$ and one from $p_{\leftarrow}$. For convenience, we use
$LM_{\rightarrow}(w_1 w_2 ... w_{i-1} \underline{\hspace{1em}})$ to denote the
distribution $p_{\rightarrow}(w_i|w_1,...,w_{i-1})$ and
$LM_{\leftarrow}(\underline{\hspace{1em}} w_{i+1} w_{i+2} ... w_n)$ to denote
$p_{\leftarrow}(w_i|w_n,...,w_{i+1})$.

\begin{table*}[t!]
\centering
\begin{tabular}{| l | l | l | l |}
	\hline
 \multicolumn{2}{|c|}{Context Only} & \multicolumn{2}{|c|}{Symmetric Pattern}\\
 \hline
 Forward dist. & Backward dist. & Forward dist. & Backward dist. \\
 \hline \hline
 \multicolumn{4}{|c|}{This is a \textbf{\textit{sound}} idea, I like it.} \\
  \hline
sad 0.02 & bad 0.12 & welcome 0.09 & funny 0.10 \\
great 0.02 & good 0.09 & practical 0.03 & beautiful 0.05 \\
huge 0.02 & great 0.06 & comprehensive 0.03 & fun 0.04 \\
very 0.02 & wonderful 0.05 & light 0.02 & simple 0.04 \\
lesson 0.02 & nice 0.04 & balanced 0.02 & interesting 0.03 \\

 \hline\hline
 \multicolumn{4}{|c|}{I liked the \textbf{\textit{sound}} of the harpsichord} \\
  \hline
idea 0.12 & sounds 0.04 & feel 0.15 & sight 0.16 \\
fact 0.07 & version 0.03 & felt 0.11 & sounds 0.11 \\
article 0.05 & rhythm 0.03 & thought 0.07 & rhythm 0.04 \\
guy 0.04 & strings 0.03 & smell 0.06 & tone 0.03 \\
concept 0.02 & piece 0.03 & sounds 0.05 & noise 0.03 \\
 \hline
\end{tabular}
\caption{Predicted substitutes for two senses of sound, for context-only and the symmetric-pattern approaches.
  } \label{tab:sensecomp}
\end{table*}
\paragraph{Context-based substitution}
In the purely context-based setup (the one used in the AI-KU system) we
represent the target word \emph{sounds} by the two distributions:

\noindent\fbox{\parbox{\columnwidth}{
\noindent$LM_{\rightarrow}($\texttt{<s>} I liked the
\underline{\hspace{1em}}$)$\\
$LM_{\leftarrow}($\underline{\hspace{1em}} of the harpsichord \texttt{</s>} $)$
}}
The resulting top predictions from each distribution are:\\
\{\emph{idea:0.12, fact:0.07, article: 0.05, guy: 0.04, concept: 0.02}\} and\\\{\emph{sounds:0.04, version: 0.03, rhythm: 0.03, strings: 0.03, piece: 0.02}\} respectively. 

\subsection{Dynamic Symmetric Patterns}
\label{ssec:sympat}
As discussed in the introduction, conditioning solely on context is ignoring
valuable information. This is evident in the resulting word distributions.
We use the coordinative symmetric pattern \emph{X and Y} in order to produce a
substitutes vector incorporating both the word and its context.  
Concretely, we represent a target word $w_i$ by $p_{\rightarrow}(w'|w_1,...,w_{i},\texttt{and})$
and $p_{\leftarrow}(w'|w_n,...,w_i,\texttt{and})$. For our running example, this
translates to:\\
\noindent\fbox{\parbox{\columnwidth}{
\noindent$LM_{\rightarrow}($\texttt{<s>} I liked the sound and
\underline{\hspace{1em}}$)$ \\
\noindent$LM_{\leftarrow}($\underline{\hspace{1em}} and sound of the harpsichord . \texttt{</s>}$)$
}}
with resulting top words: \{\emph{feel: 0.15, felt: 0.11, thought: 0.07, smell: 0.06, sounds: 0.05}\} and\\ \{\emph{sight: 0.16, sounds: 0.11, rhythm: 0.04, tone: 0.03, noise: 0.03}\}.

The distributions predicted using the \emph{and} pattern exhibit a much nicer behavior, 
and incorporate global context (resulting in sensing related substitutes)
as well as local and syntactic information that resulting from the target word itself.
Table~\ref{tab:sensecomp} compares the context-only and symmetric-pattern substitutes for two senses of the word \emph{sound}.

\subsection{Representative Generation}
\label{ssec:sampling}
To perform fuzzy clustering, we follow AI-KU and associate each instance with $k$ representatives, but deviate in the way the representatives are generated.
Specifically, each representative is a set of size $2\ell$ , containing $\ell$ samples from the forward distribution and $\ell$ samples from the backward distribution. In the symmetric pattern case above, a plausible representative, assuming $\ell=2$, would be: \{\emph{feel, sounds, sight, rhythm}\} where two words were predicted by each side LM. In this work, we use $\ell=4$ and $k=20$.

\subsection{Sense Clustering}
\label{ssec:clustering}

After obtaining $k$ representatives for each of the $n$ word instances, we
cluster the $nk$ representatives into distinct senses and translate this
hard-clustering of representatives into a probabilistic clustering of the originating instances.

\paragraph{Hard-clustering of representatives}
Let $V$ be the vocabulary obtained from all the representatives.
We associate each representative with a sparse $|V|$ dimensional bag-of-features vector, and arrange the representatives into a $nk \times |V|$ matrix $M$ where each row corresponds to a representative. We now cluster $M$'s rows into senses. We found it is beneficial to transform the matrix using TF-IDF. Treating each representative as a document, TF-IDF reduces the weight of uninformative words shared by many representatives.
We use agglomerative clustering (cosine distance, average linkage) and induce a fixed number of clusters.\footnote{In this work, we use 7 clusters, which roughly matches the number of senses for each target word in the corpus. Dynamically selecting the number of clusters is left for future work. The effect of changing the number of clusters is explored in the supplementary material.} 
We use \texttt{sklearn} \cite{scikit-learn} for both TF-IDF weighting and clustering. 

\paragraph{Inducing soft clustering over instances} 
After clustering the representatives, we induce a soft-clustering over the instances by associating each instance $j$ to sense $i$ based on the proportion of representatives of $j$ that are assigned to cluster $i$.

\subsection{Additional Processing}
\label{ssec:lemmatization}
\paragraph{Lemmatization}
The WSI task is defined over lemmas, and some target words have
morphological variability within a sense. This is especially common with verb tenses, e.g., ``I \textbf{booked} a flight'' and ``I am \textbf{booking} a flight''. As the conjunctive symmetric pattern favors morphologically-similar words, the resulting substitute vectors for these two sentences will differ, each of them agreeing with the tense of its source instance.  To deal with this, we lemmatize the predictions made by the language model prior to adding them to the representatives. Such removal of morphological inflection is straightforward when using the word distributions but much less trivial when using raw LM state vectors, further motivating our choice of working with the word distributions. The substantial importance of the lemmatization is explored in the ablation experiments in the next section, as well as in the supplementary material.
\\\noindent\textbf{Distribution cutoff and bias} Low ranked LM prediction tend to become noisier. We thus consider only the top 50 word predicted by each LM, re-normalizing their probabilities to sum to one. Additionally, we ignore the final bias vector during prediction (words are predicted via $softmax(Wx)$ rather than $softmax(Wx+b)$). This removes unconditionally probable (frequent) words from the top LM predictions.

\begin{table*}[t!]
\centering
\begin{tabular}{| l | l | l | l |}
	\hline
  Model & FNMI & FBC & AVG \\
  \hline
  \multicolumn{4}{|c|}{Original task dataset} \\
  \hline
  Ours & {\bf 11.26} $\pm$ 0.48 & 57.49 $\pm$ 0.23 & {\bf 25.43} $\pm$ 0.48 \\
  MCC-S\ & 7.62 & 55.6 & 20.58 \\
  Sense-Topic (SW)& 7.14 & 55.4 & 19.89 \\
  Sense-Topic  & 6.96 & 53.5 & 19.30 \\
  AI-KU  & 6.5 & 39.0 & 15.92 \\
  unimelb  & 6.0 & 48.3 & 17.02 \\
  \hline 
  \multicolumn{4}{|c|}{With data enrichment} \\
  \hline
  Sense-Topic (AAC) & 9.39 & {\bf 59.1 } & 23.56 \\
  Sense-Topic (AUC) & 9.74 & 54.5 & 23.04 \\
  \hline
\end{tabular}
\caption{Evaluation Results on the SemEval 2013 Task 13 Dataset. SW: Embeddings similarity based feature weighting. AAC: Extending instance sentences from their traced source. AUC: Adding similar sentences from the dataset originating corpus. We report our mean scores over 30 runs $\pm$ standard deviation}
   \label{tab:results}
\end{table*}

\section{Experiments and Results}
We evaluate our method on the SemEval 2013 Task 13 dataset \citep{semeval2013wsi}, containing 50 ambiguous words each with roughly 100 in-sentence instances, where each instance is soft-labeled with one or more WordNet senses.\\[0.1em]
\noindent\textbf{Experiment Protocol} Due to the stochastic nature of the algorithm, we repeat each experiment 30 times and report the mean scores together with the standard deviation. \\[0.1em]
\noindent\textbf{Evaluation metrics } We follow previous work \cite{wang2015sense,komninos2016structured} and evaluate on two measures: \emph{\textbf{Fuzzy Normalized Mutual Information (FNMI)}} and \emph{\textbf{Fuzzy B-Cubed (FBC)}} as well as their geometric mean (\textbf{AVG}).

\vspace{-10pt}
\paragraph{Systems} We compare against three graphical-model based systems which, as far as we know, represent the current state of the art: \textbf{MCC-S} \cite{komninos2016structured}, \textbf{Sense-Topic} \cite{wang2015sense} and \textbf{unimelb} \cite{lau2013unimelb}. We also compare against the \textbf{AI-KU} system. Wang et al. also present a method for dataset enrichment that boosted their model performance. We didn't use the suggested methods and compare ourselves to the vanilla settings, but report the enrichment numbers as well.

\paragraph{Results} Table~\ref{tab:results} summarizes the results. Our system using symmetric patterns outperforms all other setups with an AVG score of 25.4, establishing a new state-of-the-art on the task.

\begin{figure}[th!]
\centering
       \includegraphics[width=\linewidth]{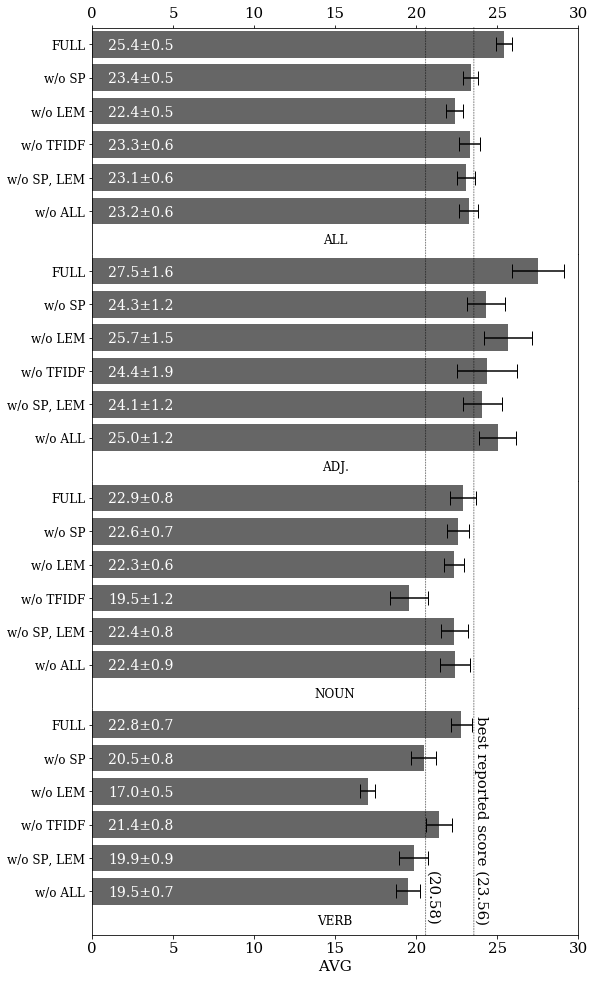}
    \caption{Ablation break down by part of speech, each part of speech was averaged across run. Bars are mean of means and error bars are standard deviations.}
    \label{fig:ablationpos}
\end{figure}

\paragraph{Ablation and analysis} We perform ablations to explore the contribution of the different components (Symmetric Patterns (SP), Lemmatization (LEM) and TF-IDF re-weighting). Figure (\ref{fig:ablationpos}) shows the results for the entire dataset (ALL, top), as well as broken-down by part-of-speech. All components are beneficial and are needed for obtaining the best performance in all cases. However, their relative importance differs across parts-of-speech. Adjectives gain the most from the use of the dynamic symmetric patterns, while nouns gain the least. For verbs, the lemmatization is crucial for obtaining good performance, especially when symmetric patterns are used: using symmetric patterns without lemmatization, the mean score drops to 17.0. Lemmatization without symmetric patterns achieves a higher mean score of 20.5, while using both yields 22.8. Finally, for nouns it is the TF-IDF scoring that plays the biggest role.

\section{Conclusions}
We describe a simple and effective WSI method based on a neural biLM and a novel dynamic application of the \emph{X and Y} symmetric pattern. The method substantially improves on the
state-of-the-art. 
Our results provide further validation that RNN-based language models
contain valuable semantic information.

The main novelty in our proposal is querying the neural LM in a creative way, in what we call \emph{dynamic symmetric patterns}. We believe that the use of such dynamic symmetric patterns (or more generally \emph{dynamic Hearst patterns}) will be beneficial to NLP tasks beyond WSI.

In contrast to previous work, we used discrete predicted word distributions rather than the continuous RNN states. This paid off by allowing us to inspect and debug the representation, as well to control it in a meaningful way by injecting linguistic knowledge in the form of lemmatization, and by distributional cutoff and TF-IDF re-weighing.  We encourage others to consider using explicit, discrete representations when appropriate.

\paragraph{Acknowledgments}
The work was supported in part by the Israeli Science Foundation (grant number 1555/15 and the German Research Foundation via the
German-Israeli Project Cooperation (DIP, grant DA 1600/1-1).

\bibliography{emnlp2018}
\bibliographystyle{acl_natbib_nourl}

\clearpage

\begin{appendices}
\section*{Supplementary Material}

\subsection*{Statistics of the SemEval 2013 Task 13 Dataset}
SemEval 2013 Task 13 consists of 50 targets, each has a lemma and a part of speech (20 verbs, 20 nouns and 10 adjectives). 
We use the dataset only for evaluation.
Most targets have around 100 labeled instances (sentences containing a usage of the target in its designated part of speech together with one or more WordNet senses assigned by human labeler). 
Exceptions are the targets of trace.n and book.v which have 37 and 22 labeled instances accordingly.
Leaving out the two anomalous targets mentioned above we are left with 4605 instances from 48 targets: 19 verb, 19 noun and 10 adjective targets. We note that the small size of the dataset should make one cautious to draw quick conclusions, yet, our results seem to be consistent.

\subsection*{Effect of the Choice of Number of Clusters}
\begin{figure}[th!]
\centering
       \includegraphics[totalheight=7.5cm]{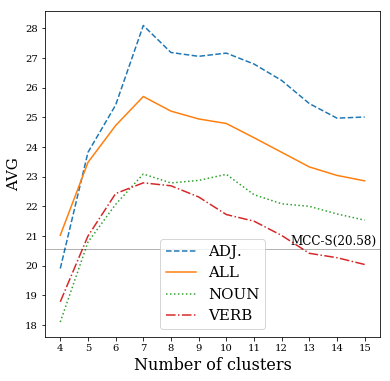}
    \caption{AVG score by number of clusters. }
    \label{fig:nclusters}
\end{figure}

An important statistic of the dataset is the number of senses per target. The average number of senses per target in the dataset is 6.94 (stdev:2.71). Breaking down by part of speech, the means and standard deviations of target senses are:  verbs: 5.90 ($\pm$1.37), nouns: 7.32 ($\pm$2.21), adjectives: 7.11 ($\pm$3.54). In this work we follow this statistic and always look for 7 clusters. Figure \ref{fig:nclusters} shows the accuracy as a function of the number of clusters. While 7 clusters indeed produces  the highest scores, all numbers in the range 4 to 15 produce state-of-the-art results. 
We leave the selection of per-instance number of clusters to future work.

Figure \ref{fig:nclusters} also tells us our system is better at inducing senses for adjectives, at least according to task score.

\subsection*{The Importance of Lemmatization}

The ablation results in the paper indicate that for verbs, using symmetric patterns without lemmatization yields poor results. We present the analysis the motivated our use of lemmatization. 
Consider the samples from the biLM with and without symmetric patterns, for the instance \emph{It was when I was a high-school student that I \textbf{became} convinced of this fact for the first time.}

\begin{tabular}{ll}
\hline
fw LM, no SP:& didn, write, 'd, learnt, start \\
bw LM, no SP:& seem, be, grow, be, be \\
\hline
fw LM, with SP:& went, got, started, wasn, \\ & loved \\
bw LM, with SP:& 1990s, decade, 1980s, \\ & afterwards, changed \\
\hline
\end{tabular}
\\
Another sentence, in another tense:
\textit{The issue will \textbf{become} more pressing as an estimated 40,000 to 50,000 Chinese, mostly unskilled, come to settle each year.}

\begin{tabular}{ll}
\hline
fw LM, no SP:& be, be, remain, likely, be \\
bw LM, no SP:& becoming, grown \\ & becoming, much, becomes \\
\hline
fw LM, with SP:& remains, remain, which, \\ & continue, how \\
bw LM, with SP:& rising, overseas, booming, \\ & abroad, expanded \\
\hline
\end{tabular}

When using the symmetric patterns, the predicted verbs tend to share the tense of the target word. 

This results in targets of different tenses having nearly distinct distributions, even when the targets share the same sense, splitting the single sense cluster to two (or more) tense clusters. We quantify this intuition by computing the correlation between tense and induced clusters (senses), as given by the Normalized Mutual Information (NMI).  We measure NMI between verb instance tense in sentence and their most probable induced cluster in the different settings, as well as the NMI of the verb instances and the gold clusters.  Table \ref{tab:tensenmi} summarize the results.
We see that in the gold clusters there is indeed very little correlation (0.15) between the the tense and the sense. When using SP but not lemmatization (w/o LEM), the correlation is substantially higher (0.67). When not using either lemmatization of SP (w/o LEM and SP) the correlation is 0.27, much closer to the gold one. Performing explicit lemmatization naturally reduces the correlation with tense, and using the full model (Final model) results in a correlation to 0.22, close to the gold number of 0.15.

\subsection*{Some Failure Modes of Dynamic Symmetric Patterns}
While the use of dynamic symmetric patterns improves performance and generally produces good substitutes for contextualized words, we also identify some failure modes and unexpected behavior.

\paragraph{Common phrases involving conjunctions}
Some target words have a strong prior to appear in common phrases involving a conjunction, causing the strong local pattern to override context-based hints. For example, when the LM is asked to complete \emph{... state and \underline{\hspace{1em}}}, its prior on \emph{church} makes it a very probable completion, regardless of context and sense. This phenomena motivated our use TF-IDF for weighing of too common words.
Relatedly, a common completion for symmetric patterns is the word \emph{then}, as \emph{and then} is a very common phrase. This completion even ignores the target word and could be troublesome if a global, cross-lemma, clustering is attempted.

\paragraph{Multi word phrases substitutes}
Sometime the LM does interpret the \emph{and} as a trigger for a symmetric relation, but on a chunk extending beyond the target word. For example, when presented with the query \emph{The human heart not only makes heart \textit{sounds} and \underline{\hspace{1em}}}, the forward LM predicted in its top twenty suggestions the word \emph{muscle}, followed by a next-word prediction of \emph{movements}. That is, the symmetry extends beyond ``sounds'' to the phrase ``heart sounds'' which could be substitutes by ``muscle movements''. We didn't specifically address this in the current work, but note that restricting the prediction to agree with the target word on part-of-speech and plurality may help in mitigating this. Furthermore, this suggests an exciting direction for moving from single words towards handling of multi-word units.

\begin{table}[t!]
\centering
\begin{tabular}{| l | l | l | l |}
\hline
Settings & NMI (mean $\pm$ STD) \\ 
\hline 
  Gold labels & 0.15 $\pm$ 0.07  \\
  \hline
  Final model & 0.22 $\pm$ 0.12\\
  w/o SP & 0.19 $\pm$ 0.08 \\
  w/o TFIDF & 0.18 $\pm$ 0.07\\
  \textbf{w/o LEM} & \textbf{0.67$\pm$ 0.12 } \\
  w/o LEM and SP & 0.26 $\pm$ 0.09  \\
  w/o ALL & 0.24 $\pm$ 0.08 \\
  \hline
\end{tabular}

\caption{Correlation between tense and sense. NMI is averaged on all verbs, using best matching sense.
SP: Symmetric Patterns, LEM: Lemmatizing predictions, ALL: LEM, SP, TFIDF.
The bold line show symmetric patterns without lemmatization excessively correlates tense and sense and provides additional validation to our hypothesis, suggesting its essential to lemmatizate when symmetric patterns are used.
  }\label{tab:tensenmi}
\end{table}

\end{appendices}

\end{document}